\RequirePackage{fix-cm}
\documentclass{article} 
\usepackage{iclr2027_conference,times}

\usepackage{amsmath,amsfonts,bm}

\def\eqref#1{equation~\ref{#1}}

\def\1{\bm{1}}

\DeclareMathAlphabet{\mathsfit}{\encodingdefault}{\sfdefault}{m}{sl}
\SetMathAlphabet{\mathsfit}{bold}{\encodingdefault}{\sfdefault}{bx}{n}

\usepackage{hyperref}
\usepackage{url}
\usepackage{amsmath,amssymb}
\usepackage{graphicx}
\usepackage{booktabs}
\usepackage{hyperref}
\usepackage{url}
\usepackage{makecell}
\usepackage{float}
\usepackage{booktabs}
\usepackage{makecell}
\usepackage{fvextra}
\usepackage{listings}
\usepackage{xcolor}
\usepackage{titletoc}
\usepackage[T1]{fontenc}

  \lstdefinestyle{prompt}{
    basicstyle=\ttfamily\scriptsize,
    breaklines=true,
    breakatwhitespace=false,
    columns=fullflexible,
    keepspaces=true,
    showstringspaces=false,
    upquote=true,
    frame=single,
    framerule=0.4pt,
    rulecolor=\color{black!35},
    backgroundcolor=\color{black!2},
    xleftmargin=5pt,
    xrightmargin=5pt,
    framexleftmargin=5pt,
    framexrightmargin=5pt,
    aboveskip=8pt,
    belowskip=8pt,
    captionpos=b
  }

\title{Continuous Context Management }

\newcommand{\authorblock}[3]{%
  \begin{tabular}[t]{c}
    {\bfseries #1}\\
    {\normalfont\mdseries #2}\\
    {\normalfont\mdseries\small\ttfamily #3}
  \end{tabular}%
}

\author{%
    \rule{0pt}{2.5em}\\[-1.5em]
  \begin{tabular}{@{}c@{\hspace{10.5em}}c@{}}
    \authorblock{William Hoy}
      {University of Miami}
      {wjh58@miami.edu}
    &
    \authorblock{Jingxuan Fan}
      {Harvard University}
      {jfan@g.harvard.edu}
    \\[3.5em]
    \authorblock{Nurcin Celik}
      {University of Miami}
      {celik@miami.edu}
    &
    \authorblock{Xu Pan}
      {Harvard University}
      {xupan@fas.harvard.edu}
  \end{tabular}%
}

\iclrfinalcopy 
\begin{document}

\maketitle

\begin{abstract}
  Long-horizon large language model (LLM) agents commonly retain their complete
  interaction history until compaction is triggered at a predefined threshold.
  We study Continuous Context Management (CCM), which performs compaction at
  every turn to prevent interaction history from accumulating in the active
  prompt. At each turn, a CCM agent emits an updated memory together with an
  environment action; its next prompt contains the original task, retained
  memory, and newest observation rather than the complete transcript. We first
  evaluate CCM without fine-tuning on TerminalBench-2 using Claude Sonnet 4.6,
  Claude Opus 4.6, GLM-5, and Kimi K3. CCM substantially reduces cumulative
  input usage and active-prompt size, although it lowers task success for most
  models while preserving performance for Kimi K3. We use GRPO with privileged full-history distillation
  to improve CCM in open-weight models. A frozen copy
  of the student's initial model scores each sampled student action under the complete
  history reconstructed from that student's rollout, providing dense
  action-token supervision without a separate teacher rollout or reference
  solution. On WebShop, this objective substantially improves CCM over GRPO at
  both evaluated model scales and surpasses full-history GRPO for
  Qwen3-4B-Instruct, though not for Qwen3-8B. On Endless Terminals, the augmented
  method provides a modest
improvement over GRPO, with both CCM policies outperforming the untrained
  full-history baseline. These results demonstrate that CCM is a viable
  inference paradigm for agents operating with substantially reduced retained
  context and that its performance can be improved through reinforcement
  learning with privileged full-history distillation.
\end{abstract}

\section{Introduction}

  Long-horizon language agents must preserve useful information across repeated interactions with their environments. A shopping agent may need to remember product attributes encountered several pages earlier, while a terminal agent may need to retain the outcome of a previous command. Keeping the complete interaction history makes this information directly accessible, but causes the active context to grow as the task proceeds. The accumulated history is bounded by the model's context window, and model performance can degrade even within this limit as inputs grow longer \citep{liu2024lost,du2025context,hong2025context} or include irrelevant information \citep{shi2023large}. 
  
  A common solution is to compact the interaction history when the context reaches a predefined token threshold, replacing earlier content with a shorter summary \citep{wu2025resum,li2026compactionrl}. Repeated compaction allows a session to continue beyond the length of a single context window. However, the context still accumulates between compaction events and can occupy a large fraction of the available window before being compressed again.
  An ultimate form of this accumulate-compact paradigm is that memory management become part of every action step, with the agent continually updating the information it carries forward. We study this setting which we call Continuous Context Management (CCM), where the agent produces an updated memory alongside each action, its next context constructed from the updated memory and the newest observation. “Continuous” refers to memory management at the finest step resolution, not its meaning in mathematics.
  This is particularly useful when GPU memory is a major constraint, such as deploying LLMs locally on consumer-grade GPUs, where keeping the active context compact can reduce the token cache memory, allowing larger models to run within the same GPU memory budget.

 We first evaluate CCM as an out-of-the-box inference paradigm on
  TerminalBench-2 using Claude Sonnet 4.6, Claude Opus 4.6, GLM-5, and Kimi K3.
  Across these models, CCM substantially reduces cumulative input usage and
  typical active-prompt size. Success declines for Sonnet, Opus, and GLM-5, but
  increases slightly for Kimi K3. The resulting monetary savings depend strongly
  on provider-specific prompt-caching policies and token prices: prefix caching
  can substantially reduce the cost of full-history prompting despite its greater
  token usage. These results demonstrate CCM's potential to reduce context usage
  while revealing a performance gap for three of the four models, motivating our
  investigation of training methods for CCM agents.

  Prior work has established the feasibility of learning such memory-management behaviors, such as MEM1, MemAgent, MEMENTO, and Compaction RL \citep{zhou2025mem1, yu2025memagent, kontonis2026memento, li2026compactionrl}. Building on these approaches, particularly the per-turn consolidation used in MEM1, we combine two training signals to train agents both to manage memories and to use them to generate actions. 
  These two capabilities are closely coupled. An agent may preserve the relevant facts yet fail to use them when choosing an action; conversely, an effective action policy cannot reliably compensate for essential information omitted from its memory. Reinforcement learning can optimize both capabilities through task rewards, but episode-level outcomes provide limited guidance about individual task actions. To provide additional supervision for actions generated from compact memory, we use an additional distillation objective building on on-policy context distillation \citep{ye2026opd} and the gated self-distillation approach of SDAR \citep{lu2026sdar}, where the teacher sees the full interaction history as privileged information. The teacher is initialized with the same weights as the student before training and remains frozen throughout training. At each step, it receives the full history preceding the current action and scores the student-sampled action tokens autoregressively. A token-level teacher–student confidence-gap gate weights the policy gradient on action tokens. This procedure provides action-token supervision without a separately trained teacher, additional teacher rollout, or successful reference trajectory. 

  We evaluate this training recipe on WebShop and Endless Terminals, focusing on improving agent performance while using only a fraction of the prompt context required by full-history baselines. On WebShop, adding full-history distillation substantially improves task success over CCM trained with GRPO alone for both models. On Endless Terminals, both trained CCM policies outperform the untrained
full-history baseline while maintaining compact model-written memories.
Distillation provides a modest
improvement over CCM + GRPO. Together, these results show that our training recipe improves CCM’s task performance while retaining its substantial savings in prompt length.

Our main contributions are:
  \begin{itemize}

  \item We study Continuous Context Management (CCM), in which agents update a compact memory at every action step. Our evaluation on TerminalBench-2 demonstrates substantial reductions in input-token usage without additional training, and characterizes the task performance and API-cost tradeoffs.
  
  \item We introduce a training recipe that combines GRPO with privileged full-history distillation. Experiments on WebShop and Endless Terminals show that training improves CCM’s task performance while substantially reducing context length.

    
    


  \end{itemize}

\section{Related Works}
\label{sec:related-work}

  \paragraph{Context management for long-horizon agents.}
  ReAct-style agents retain the full sequence of reasoning, actions, and
  observations in subsequent prompts~\citep{yao2022react}, so the context grows
  with the horizon and performance can degrade on long
  inputs~\citep{du2025context}. Inference-time methods extend the horizon through
  external memory, periodic summarization, structured compression, and evolving
  context representations~\citep{packer2023memgpt,wu2025resum,kang2025acon,
  wan2025compass,zhang2026ace,li2026selfcompact}, and several works train models
  to generate summaries or decide when to
  compress~\citep{lu2025supo,li2026acm,zhang2026autocompact,yu2025memagent,
  kontonis2026memento}. MEM1~\citep{zhou2025mem1} is the closest in formulation to our method. At every
  turn the agent emits an internal state that consolidates its previous state
  with the newest observation, and earlier turns are pruned. MEM1 learns this behavior from outcome rewards
  with PPO, training on the concatenated trajectory
  with an attention mask that restricts each token to the context available
  when it was generated. Our study differs from MEM1 in several respects. First, we show that frontier models can use CCM out-of-box without additional training, substantially reducing context usage, although most evaluated models incur a drop in task performance. Second, rather than relying on outcome rewards
  alone to train the model to manage its state, we combine episode-level GRPO with a designed token-level distillation objective. 
  

  \paragraph{Reinforcement learning and on-policy distillation.}
  Reinforcement learning methods such as GRPO provide trajectory-level
  supervision from environment or verifier rewards~\citep{shao2024deepseekmath,
  dong2025,feng2025}. On-policy distillation, including on-policy
  self-distillation, complements this feedback with dense token-level guidance
  from a stronger teacher or from the same policy conditioned on privileged
  information~\citep{ye2026opd,yang2026opd,team2026opd,glm5team2026opd,
  zhao2026opsd,he2026opsd,zhang2026opsd}. \citet{lu2026sdar} proposed Self-Distilled Agentic Reinforcement
  Learning (SDAR), which combines reinforcement learning and on-policy
  self-distillation using token-level gates based on student uncertainty or the
  teacher--student confidence gap. We
  adapt its confidence-gap gating to multi-turn CCM agents. For each sampled
  action token, we compare its probability under the current policy's compact
  CCM context with its probability under a frozen copy of the student's
  initial model, conditioned on the reconstructed full history. The teacher
  weights remain fixed throughout training. This provides an auxiliary
  action-token distillation signal without requiring a separately trained
  teacher, an additional teacher rollout, or a successful reference trajectory.

\section{Continuous Context Management}
\label{sec:ccm}

\subsection{Method Description}

  For a task $x$, let $o_0$ denote the initial environment observation and
  $m_0$ the initially empty memory. At turn $n\geq 1$, a Continuous Context
  Management (CCM) agent receives observation $o_{n-1}$ and retained memory
  $m_{n-1}$, and generates updated memory $m_n$ and action $a_n$ according to
  \begin{equation}
    (m_n, a_n)
    \sim
    \pi_\theta\!\left(
      \cdot \mid x,\, m_{n-1},\, o_{n-1}
    \right),
    \label{eq:ccm-sample}
  \end{equation}
  The generated response consists of an updated memory followed by an environment action, \(y_n = [\,m_n\,;\,a_n\,]\).
  Executing $a_n$ produces the next observation $o_n$. The agent then
  constructs the prompt for turn $n+1$ from $x$, $m_n$, and $o_n$. Previous responses and observations are not
  included directly. Information from earlier turns therefore remains available
  only if the agent decides to preserve it in $m_n$.

  \begin{figure}[H]
    \centering
    \includegraphics[width=0.855\linewidth]
      {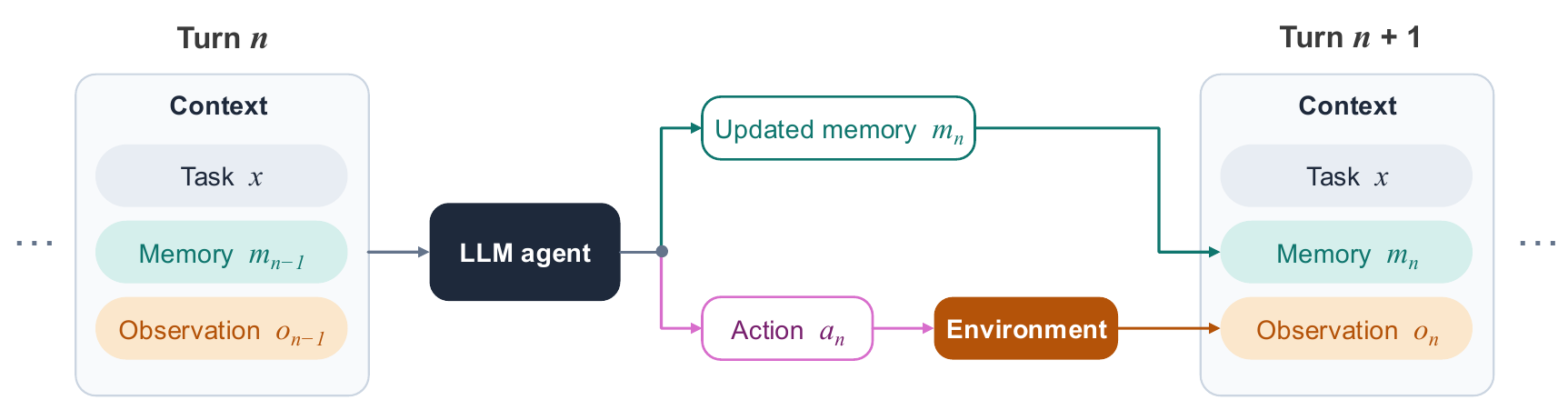}
    \caption{Continuous Context Management. At each turn, the agent
    receives the original task, its memory from the preceding turn, and the
    current observation. It emits an updated memory followed by an environment
    action. The next prompt is constructed from the updated memory and newest
    observation rather than the complete interaction transcript.}
    \label{fig:ccm-rollout}
  \end{figure}

\subsection{TerminalBench-2 results}
  \label{sec:results:tb2}

  Terminal-Bench 2.0~\citep{merrill2026terminal} contains 89 challenging,
  human-curated tasks that require agents to complete realistic workflows through
  terminal interaction. We evaluate CCM on all 89 tasks using Claude Sonnet 4.6,
  Claude Opus 4.6, GLM-5, and Kimi K3. For each model, we compare CCM against a
  standard full-history baseline using pass@1.

  Table~\ref{tab:tb2-main} reports task success, token usage,
  and estimated API cost per task; Table~\ref{tab:tb2-four-models-full}
  additionally reports interaction length and detailed token accounting.
  Cost estimates for Claude Sonnet 4.6, Claude Opus 4.6, and Kimi K3 account
  for provider-side prompt caching. GLM-5 is reported without prompt caching
  because Amazon Bedrock does not support it for this model. Inference and
  token-accounting configurations are provided in Appendix~\ref{app:tb2-setup}.

\begin{table*}[t]
  \centering
  \caption{Terminal-Bench 2.0 pass@1 results over 89 tasks. Costs are estimated
means per task and include prompt caching where available; prompt caching was
unavailable for GLM-5 through Amazon Bedrock.}
  \label{tab:tb2-main}
  \vskip 10pt
  \fontsize{8.5}{9.5}\selectfont
  \setlength{\tabcolsep}{4pt}
  \begin{tabular*}{\textwidth}{
    @{\extracolsep{\fill}}llccccc@{}
  }
    \toprule
    & & &
    \multicolumn{2}{c}{Peak prompt tokens}
    & & \\
    \cmidrule(lr){4-5}
    Model & Mode
      & \makecell[c]{Success,\\$n$ (\%)}
      & Median
      & Maximum
      & \makecell[c]{Maximum retained\\memory}
      & \makecell[c]{Estimated cost\\per task} \\
    \midrule

    Claude Sonnet 4.6
      & Baseline & \textbf{38 (42.7)}
      & 16{,}673 & 131{,}157 & -- & \textbf{\$0.468} \\
    {}
      & CCM & 28 (31.5)
      & \textbf{3{,}568} & \textbf{50{,}614} & 1{,}219 & \$0.472 \\

    \addlinespace
    Claude Opus 4.6
      & Baseline & \textbf{52 (58.4)}
      & 15{,}405 & 103{,}512 & -- & \textbf{\$0.831} \\
    {}
      & CCM & 36 (40.4)
      & \textbf{4{,}280} & \textbf{51{,}457} & 1{,}403 & \$1.087 \\

    \addlinespace
    GLM-5
      & Baseline & \textbf{31 (34.8)}
      & 17{,}560 & 116{,}407 & -- & \$0.515 \\
    {}
      & CCM & 26 (29.2)
      & \textbf{4{,}279} & \textbf{47{,}287} & 561 & \textbf{\$0.124} \\

    \addlinespace
    Kimi K3
      & Baseline & 56 (62.9)
      & 9{,}472 & 74{,}313 & -- & \textbf{\$0.302} \\
    {}
      & CCM & \textbf{57 (64.0)}
      & \textbf{3{,}243} & \textbf{27{,}645} & 1{,}058 & \$0.622 \\

    \bottomrule
  \end{tabular*}
\end{table*}

As shown in Table~\ref{tab:tb2-main}, CCM substantially reduces peak prompt
size across all four models. Median episode-level peaks decrease from
9{,}472--17{,}560 tokens under full-history prompting to 3{,}243--4{,}280
tokens under CCM. The largest retained CCM memories contain only
561--1{,}403 tokens. Maximum CCM prompts can nevertheless be considerably
larger because each prompt also includes the newest terminal observation,
which may contain up to 60{,}000 characters. CCM limits accumulated interaction
history but does not compress the current observation.

CCM also reduces mean cumulative input tokens per task by 83.1\% for Claude
Sonnet 4.6, 81.5\% for Claude Opus 4.6, 84.3\% for GLM-5, and 75.8\% for
Kimi K3. This compression is accompanied by lower pass@1 success for Sonnet,
Opus, and GLM-5, with decreases of 11.2, 18.0, and 5.6 percentage points,
respectively. In contrast, Kimi K3 improves slightly under CCM, from 62.9\%
to 64.0\%.

The monetary effect depends strongly on provider-side prompt caching. For
Sonnet, the two methods have nearly identical estimated costs: \$0.468 per
task for full-history prompting and \$0.472 for CCM. For Opus and Kimi,
caching makes full-history prompting less expensive despite its substantially
greater cumulative input usage: \$0.831 versus \$1.087 for Opus and \$0.302
versus \$0.622 for Kimi. Because our GLM-5 endpoint does not support prompt
caching, its input reduction lowers estimated cost from \$0.515 to \$0.124
per task. Overall, CCM consistently reduces peak and cumulative context usage,
but its effects on task success and monetary cost depend on the model and the
provider's caching and pricing policies.

\section{Improving Continuous Context Management via Reinforcement Learning}
\label{sec:method}

    \subsection{GRPO with Self-Distillation}

  We improve the CCM policy by combining GRPO and a
  distillation-based objective. We refer to the CCM policy defined in
  Section~\ref{sec:ccm} as the \emph{student}, and denote its response at turn
  $n$ by \(y^S_n := y_n = [\,m_n\,;\,a_n\,]\).
  The student rollouts follow the CCM interaction process illustrated in
  Figure~\ref{fig:ccm-rollout}. GRPO optimizes all valid tokens in the student
  response, including memory and action tokens (details in Appendix \ref{app:grpo-objective}). 


  Beyond GRPO, we also use a distillation-based objective. The \emph{teacher} policy is a frozen copy of the student model before training. Its weights remain fixed throughout training, while the student is updated. At each turn, the teacher uses the full environment history to provide action-token supervision for the student acting from compact memory.
  
  Concretely, we reconstruct the student's environment history before
  the current action $a_n$ as privileged information,
  $h^S_n = \bigl(o_0,a_1,o_1,\ldots,a_{n-1},o_{n-1}\bigr)$,
  where $o_0$ is the initial observation. This history includes only prior
  actions and observations, excluding the current action $a_n$.
For token position $t$ within the current action $a_n$, the student is
  conditioned on $s^C_{n,t}
    =
    P_C\bigl(x,m_{n-1},o_{n-1},m_n,a_{n,<t}\bigr)$,
  where $P_C$ is the CCM prompt renderer, $m_n$ has already been generated
  as the first part of the current response, and $a_{n,<t}$ denotes the
  preceding action-token prefix.
The privileged teacher is conditioned on $s^+_{n,t}
    =
    P_F\bigl(x,h^S_n,a_{n,<t}\bigr)$,
  where $P_F$ is the full-history prompt renderer. When scoring $a_{n,t}$,
  the teacher receives only $h^S_n$ and the preceding action-token prefix
  $a_{n,<t}$ in addition to the task $x$; neither the target token nor
  subsequent action tokens are included in its conditioning context. 
  The student and teacher score the same sampled action token from the same
  trajectory. The student uses its current weights and compact memory,
  whereas the teacher uses fixed initial weights and the complete
  pre-action environment history.

 \vskip 10 pt
    \begin{figure}[H]
    \centering
    \includegraphics[width=0.9025\linewidth]
      {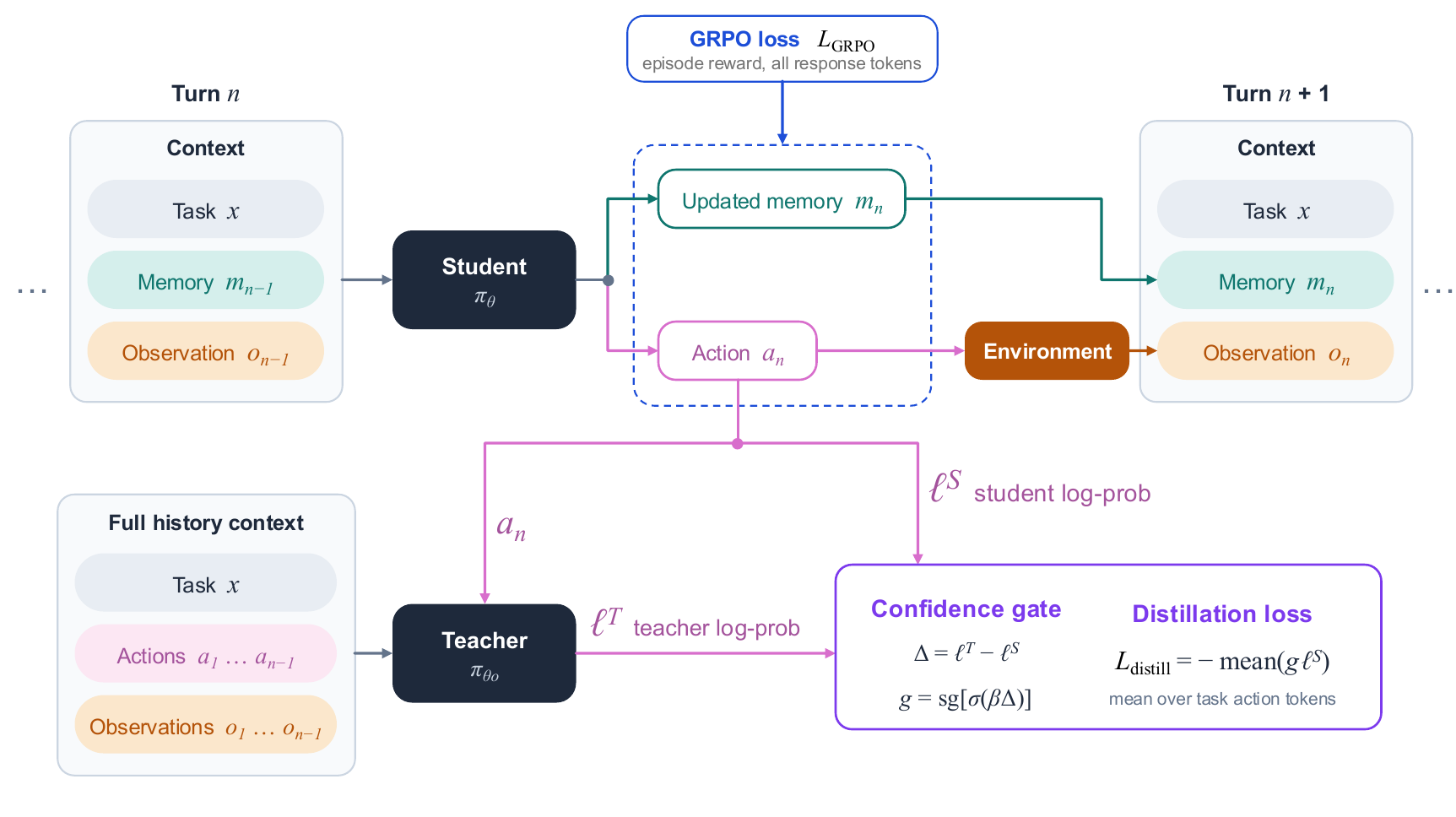}
    \caption{Training CCM agents with GRPO and full-history self-distillation.
\textbf{Top:} the student $\pi_\theta$ rolls out under CCM. GRPO with episode task reward optimizes all response tokens, both memory
and action.
\textbf{Bottom:} the teacher $\pi_{\theta_0}$ is a copy of the student's
initial model, with weights held fixed throughout training. It receives the
pre-action environment history reconstructed from the student's rollout and
scores the student-sampled action autoregressively. For each action token, the teacher log-probability $\ell^T$ and
the student log-probability $\ell^S$ give the confidence gap $\Delta$.
The gap sets the gate $g$, which weights the
distillation loss $\mathcal{L}_{\text{distill}} = -\,\mathrm{mean}(g\,\ell^S)$ over valid
action tokens only. The total training loss is composed of the GRPO loss and distillation loss.}
    \label{fig:teacher-scoring}
  \end{figure}
  


  For a student sampled action token $a_{n,t}$, we define the teacher student
  confidence gap as
  \begin{equation}
    \Delta_{n,t}
    =
    \operatorname{stopgrad}\!\left[
      \log \pi_{\theta_0}\bigl(a_{n,t}\mid s^+_{n,t}\bigr)
      -
      \log \pi_\theta\bigl(a_{n,t}\mid s^C_{n,t}\bigr)
    \right],
    \label{eq:gap}
  \end{equation}
  where $\theta_0$ denotes the student's initial parameters before any
  training updates. The teacher $\pi_{\theta_0}$ remains frozen throughout
  training and is never refreshed from the updated student. The operator
  $\operatorname{stopgrad}[\cdot]$ treats its argument as constant during
  backpropagation.

  Following the gap-gating formulation of SDAR~\citep{lu2026sdar}, we convert the confidence gap into a token-level gate, $g_{n,t}
    =
    \operatorname{stopgrad}\!\left[
      \sigma\bigl(\beta\Delta_{n,t}\bigr)
    \right]$,
  where $\sigma$ is the logistic sigmoid and $\beta>0$ controls gate
  sharpness. Tokens receiving greater confidence from the privileged
  teacher have $\Delta_{n,t}>0$ and receive larger gating weights.
  Tokens already receiving greater confidence from the student have
  $\Delta_{n,t}<0$ and are attenuated, but retain a small positive weight.


  Let $M_{i,n,t}\in\{0,1\}$ indicate that token $t$ belongs to a valid action
  block at turn $n$ of rollout $i$ and has valid student and teacher scores.
  The distillation loss is averaged over all eligible action tokens:
  \begin{equation}
    L_{\mathrm{distill}}
    =
    {
      -\displaystyle\sum_{i,n,t}
      M_{i,n,t}g_{i,n,t}
        \log \pi_\theta
          \bigl(a_{i,n,t}\mid s^C_{i,n,t}\bigr)
    }/{
      \displaystyle\sum_{i,n,t}M_{i,n,t}
    }.
    \label{eq:distillation-loss}
  \end{equation}


  The distillation loss applies to all valid student-generated action tokens (task actions only, unlike GRPO which is on both memory and task actions). It weights each token's gradient according to the teacher–student log-probability gap. When the full-history teacher assigns a token higher probability than the compact-context student, the token receives a larger change of gating weight, encouraging the student to reproduce the token supported by the additional context. Negative-gap tokens receive smaller gating weights, scaling down the magnitude of gradients. In Appendix \ref{app:SDAR}, we show at fixed contexts under on-policy sampling, the expected update is equivalent to using a bounded, centered sigmoid transformation of OPD’s signed log-probability gap, recovering a scaled OPD gradient locally for small gaps.


  The combined
  training loss is $L_{\mathrm{total}}
    =
    L_{\mathrm{GRPO}}
    +
    \lambda_{\mathrm{distill}}L_{\mathrm{distill}}$.
  We set
  $\lambda_{\mathrm{distill}}=0.01$ and $\beta=5$. Other training config details are shown in Appendix \ref{app:experimental-details}.


\subsection{WebShop Results}
\label{sec:results:webshop}

\begin{figure}[H]
    \centering
    \includegraphics[width=0.95\linewidth]
      {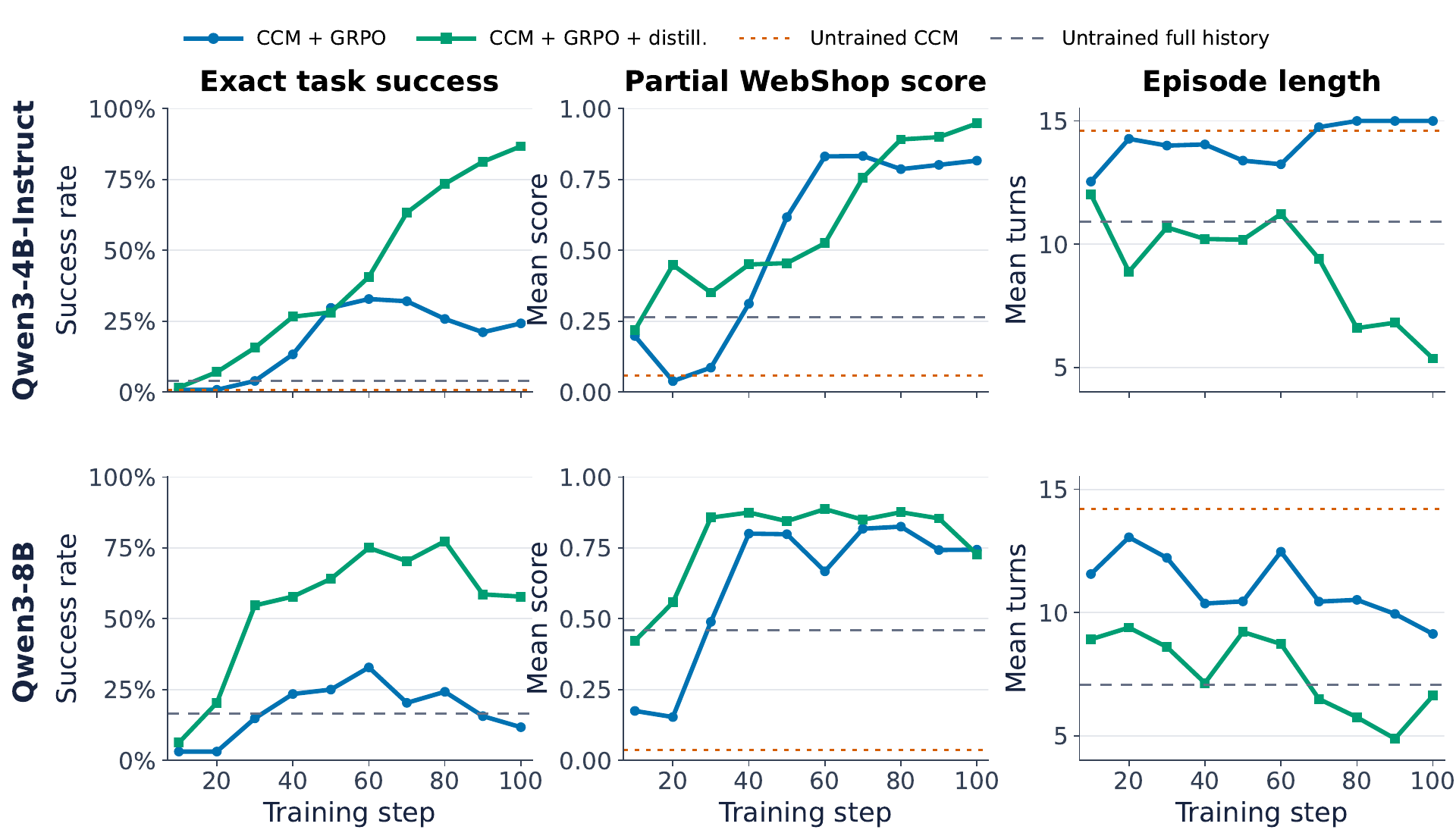}
    \caption{WebShop checkpoint evaluation for Qwen3-4B-Instruct (top)
    and Qwen3-8B (bottom) on a fixed 128-task pass@1 evaluation with
    temperature 0.4, seed 42, and a 15-turn limit. Columns report exact
    task success, partial WebShop score, and mean episode length. Solid
    curves show CCM + GRPO and CCM + GRPO + distillation across
    checkpoints 10--100. Dashed lines indicate the corresponding
    untrained CCM and full-history performance.}
    \label{fig:webshop}
  \end{figure}

WebShop~\citep{yao2022webshop} is a simulated shopping benchmark that tests
an agent's ability to navigate webpages, select product options, and complete
purchases. Figure~\ref{fig:webshop} compares CCM + GRPO with CCM + GRPO +
distillation across training checkpoints for Qwen3-4B-Instruct and Qwen3-8B.
The untrained CCM and full-history conditions provide fixed reference points.
Table~\ref{tab:prefix-cache} additionally reports full-history GRPO and
summarizes task performance, memory behavior, and prompt size at each trained
method's best observed checkpoint.

  At step 100, adding privileged distillation substantially improves CCM
  over CCM + GRPO at both model scales. For Qwen3-4B-Instruct, adding privileged
  distillation improves exact success by 62.50 percentage points, from
  24.22\% to 86.72\%, while partial score increases from 0.817 to 0.948. The distillation-augmented CCM policy also exceeds
  full-history GRPO, which achieves 75.78\% exact success and a partial score
  of 0.839. It completes tasks in 5.36 turns on average, compared with
  5.85 turns for full-history GRPO and 15.00 turns for CCM + GRPO.

For Qwen3-8B at step 100, adding privileged distillation improves exact
  success by 46.09 percentage points over CCM + GRPO, from 11.72\% to
  57.81\%. Its partial score is slightly lower, at 0.728 versus 0.744. Full-history GRPO remains the strongest condition for this
  model, reaching 80.47\% exact success and a partial score of 0.907.
  Mean episode length is 6.63 turns for CCM + GRPO + distillation,
  9.13 turns for CCM + GRPO, and 5.55 turns for full-history GRPO.

  Condensed matched trajectories are reported in
  Appendix~\ref{app:qualitative-webshop}. In the selected task, the
  untrained policies fail during search or navigation, while CCM + GRPO
  reaches the correct product but repeatedly selects an already chosen
  option. CCM + GRPO + distillation instead follows the required
  product--color--size--purchase sequence.

\begin{table*}[t]
  \centering
  \caption{Best observed task success and context behavior on the fixed
  128-task WebShop evaluation. Prompts are truncated at the 4{,}096-token
  model-input limit.}
  \label{tab:prefix-cache}
  \vskip 10pt

  \fontsize{8.5}{9.5}\selectfont
  \setlength{\tabcolsep}{1.5pt}
  \begin{tabular*}{\textwidth}{
    @{\extracolsep{\fill}}lcccccccc@{}
  }
    \toprule
    &
    \multicolumn{2}{c}{Best performance}
    & \multicolumn{2}{c}{Memory behavior}
    & \multicolumn{2}{c}{Memory size}
    & \multicolumn{2}{c}{Peak prompt tokens} \\
    \cmidrule(lr){2-3}
    \cmidrule(lr){4-5}
    \cmidrule(lr){6-7}
    \cmidrule(lr){8-9}
    Condition
      & \makecell[c]{Exact\\success (\%)}
      & Step
      & \makecell[c]{Update\\rate (\%)}
      & \makecell[c]{Prefix\\retention (\%)}
      & Mean
      & Maximum
      & Median
      & Maximum \\
    \midrule

    \multicolumn{9}{@{}l}{\emph{Qwen3-4B-Instruct}} \\
    Untrained full history
      & 3.91 & --
      & -- & -- & -- & --
      & 4{,}096 & 4{,}096 \\
    Full-history GRPO
      & 81.25 & 70
      & -- & -- & -- & --
      & 1{,}721 & 3{,}076 \\
    Untrained CCM
      & 0.78 & --
      & 57.6 & 69.6 & 225.7 & 861
      & 1{,}232 & 1{,}908 \\
    CCM + GRPO
      & 32.81 & 60
      & 12.9 & 98.5 & 95.6 & 147
      & 1{,}045 & 1{,}327 \\
    CCM + GRPO + distillation
      & \textbf{86.72} & 100
      & 100.0 & 77.2 & 60.1 & 89
      & \textbf{1{,}041} & \textbf{1{,}279} \\

    \addlinespace
    \multicolumn{9}{@{}l}{\emph{Qwen3-8B}} \\
    Untrained full history
      & 16.41 & --
      & -- & -- & -- & --
      & 2{,}258 & 4{,}096 \\
    Full-history GRPO
      & \textbf{82.81} & 70
      & -- & -- & -- & --
      & 1{,}738 & 4{,}096 \\
    Untrained CCM
      & 0.00 & --
      & 15.0 & 96.7 & 511.8 & 1{,}013
      & 1{,}582 & 2{,}127 \\
    CCM + GRPO
      & 32.81 & 60
      & 9.1 & 99.4 & 510.6 & 994
      & 1{,}200 & 1{,}954 \\
    CCM + GRPO + distillation
      & 77.34 & 80
      & 96.9 & 82.3 & 185.3 & 421
      & \textbf{1{,}149} & \textbf{1{,}435} \\

    \bottomrule
  \end{tabular*}
\end{table*}

 Table~\ref{tab:prefix-cache} reports memory-update behavior and prefix reuse at each trained method's best observed checkpoint. The CCM + GRPO policies modify their memories infrequently.
  Qwen3-4B-Instruct and Qwen3-8B have memory-update rates of 12.9\% and
  9.1\%, while retaining nearly their entire preceding
  memory prefixes. Adding distillation produces more active
  memory management. Qwen3-4B-Instruct changes its memory on every
  consecutive turn while retaining 77.2\% of the preceding prefix;
  Qwen3-8B changes it on 96.9\% of turns while retaining 82.3\%. The distillation objective is intended to improve memory use rather than memory management directly. However, improve memory use helps the GRPO to be more effective, consequently improves both memory action and task action.


Relative to CCM + GRPO, adding privileged distillation reduces cumulative
  input from 10{,}636 to 3{,}720 tokens for Qwen3-4B-Instruct and from
  9{,}929 to 5{,}578 tokens for Qwen3-8B, reductions of 65.0\% and 43.8\%.
  These differences partly reflect earlier task completion rather than
  context compression alone. CCM also does not automatically reduce total
  input, that the untrained Qwen3-8B CCM policy takes more turns than its
  full-history counterpart and consequently processes more tokens.

\subsection{Endless Terminals Results}
\label{sec:results:endless}

\begin{figure}[]
    \centering
    \includegraphics[width=0.8075\linewidth]
      {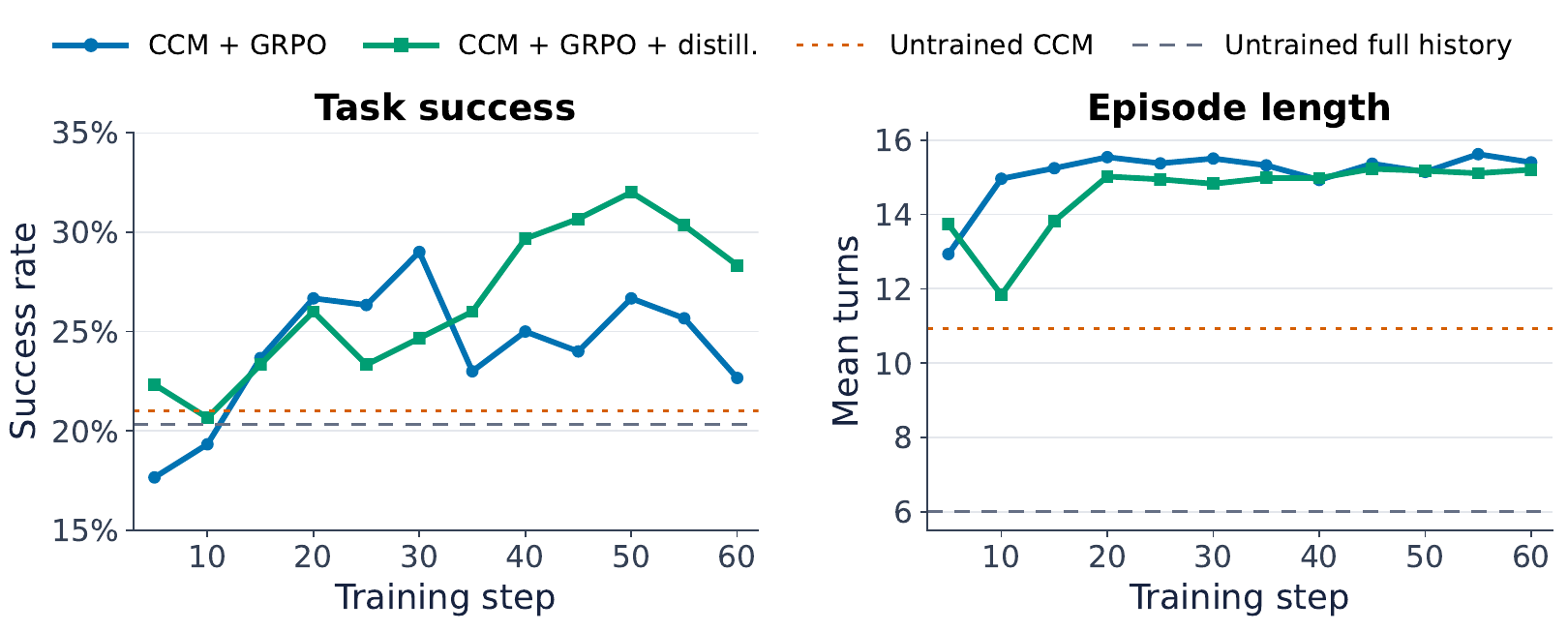}
    \caption{Endless Terminals checkpoint evaluation for Qwen3-8B on the
    fixed 300-task test set with a 16-turn limit. Curves report task
    success (left) and mean episode length (right) for CCM + GRPO and
    CCM + GRPO + distillation. Dashed lines indicate the corresponding
    untrained CCM and full-history policies.}
    \label{fig:endless-results}
  \end{figure}

We also compare CCM + GRPO against CCM + GRPO + distillation on the Endless Terminals \citep{gandhi2025endless} using Qwen3-8B, where agents execute terminal commands to manage files, process data, analyze logs, write scripts, and operate databases. The Figure~\ref{fig:endless-results}
  reports task success and mean episode length. Dashed lines show the
  untrained CCM and full-history policies under the same evaluation
  protocol.

  The untrained full-history and CCM policies perform similarly, solving
  20.33\% and 21.00\% of tasks, respectively. This difference corresponds
  to only two tasks out of 300. The two conditions solve 43 tasks in
  common, while 18 are solved only with full history and 20 only with CCM.
  One possible explanation for such a small gap in contrast with the other benchmarks is that Endless Terminals externalizes part of
  the relevant state in its persistent sandbox as files, and shell state survive across turns, allowing omitted
  information from the context to be recovered through additional inspection commands.

The best observed success is 3.00 percentage points higher with
CCM + GRPO + distillation: 32.00\% at step 50, compared with 29.00\%
at step 30 for CCM + GRPO. At step 60, the two methods achieve 22.67\% and 28.33\%,
respectively. Both selected CCM policies use nearly the full 16-turn allowance. Full-history GRPO achieves the strongest performance, reaching 35.33\% at step 30.
Although both trained CCM methods outperform the untrained full-history
and CCM policies, neither matches full-history GRPO. Examples of
condensed matched trajectories appear in
Appendix~\ref{app:qualitative-endless}. We report one task solved only by
CCM + GRPO + distillation among the two trained CCM policies and one
solved only by CCM + GRPO.

\begin{table*}[t]
  \centering
  \caption{Best observed task success and memory behavior through training
  step 60 on the fixed 300-task Endless Terminals evaluation with
  Qwen3-8B. For each trained method, we report the evaluated checkpoint
  with the highest success, breaking ties in favor of the earlier
  checkpoint.}
  \label{tab:endless-prefix-cache}
  \vskip 10pt

  \fontsize{8.5}{9.5}\selectfont
  \setlength{\tabcolsep}{4pt}
  \begin{tabular*}{\textwidth}{
    @{\extracolsep{\fill}}lcccccc@{}
  }
    \toprule
    &
    \multicolumn{2}{c}{Best performance}
    & \multicolumn{2}{c}{Memory behavior}
    & \multicolumn{2}{c}{Memory size (tokens)} \\
    \cmidrule(lr){2-3}
    \cmidrule(lr){4-5}
    \cmidrule(lr){6-7}
    Condition
      & \makecell[c]{Success\\(\%)}
      & Step
      & \makecell[c]{Update\\rate (\%)}
      & \makecell[c]{Prefix\\retention (\%)}
      & Mean
      & Maximum \\
    \midrule

    Untrained full history
      & 20.33 & --
      & -- & -- & -- & -- \\
    Full-history GRPO
      & \textbf{35.33} & 30
      & -- & -- & -- & -- \\
    Untrained CCM
      & 21.00 & --
      & 34.8 & 81.2 & 63.9 & 346 \\
    CCM + GRPO
      & 29.00 & 30
      & 16.9 & 94.6 & 77.1 & 720 \\
    CCM + GRPO + distillation
      & 32.00 & 50
      & 36.9 & 72.9 & 144.0 & 906 \\

    \bottomrule
  \end{tabular*}
\end{table*}

Table~\ref{tab:endless-prefix-cache} reports task success and memory
behavior through training. CCM + GRPO + distillation updates its
memory more frequently than CCM + GRPO, at 36.9\% versus 16.9\% of
consecutive turns, and preserves a smaller prefix of the preceding
memory, at 72.9\% versus 94.6\%. The distillation-augmented policy also
maintains a larger memory, with a mean of 144.0 tokens and a maximum of
906 tokens, compared with a mean of 77.1 tokens and a maximum of 720
tokens for CCM + GRPO. We omit peak-prompt statistics because the terminal observation can itself be very large in this benchmark, we set a cut off limit of the context length for all the conditions, and every condition hits this limit at least in one episode. Because CCM manages retained interaction history but
does not compress the newest observation, these prompt-size statistics do
not cleanly measure the amount of memory retained by the agent, but dominated by long observations and our hard cut-off limit on its length.

 \subsection{General-Capability Retention}
  \label{sec:results:retention}

  Prior work shows that on-policy distillation can improve task performance
  while mitigating forgetting of existing capabilities
  \citep{shenfeld2026selfdistillation,ye2026opd}. We investigate whether
  privileged full-history distillation provides a similar benefit beyond
  GRPO alone during agent training under CCM. To measure retention, we
  evaluate the pretrained base models and checkpoint-100/60 (WebShop/Endless Terminal) policies on
  MMLU-Pro, HellaSwag, and IFEval. Table~\ref{tab:retention-eval} reports
  mean scores over three decoding seeds.

  \begin{table*}[t]
    \centering
    \caption{General-capability retention after agent post-training. Values
are mean percentage scores over three decoding seeds, with sample standard
deviations. Parentheses report absolute percentage-point changes from the
corresponding pretrained base model, computed before rounding.}
    \label{tab:retention-eval}
    \vskip 10pt
    \fontsize{8.5}{9.5}\selectfont
    \setlength{\tabcolsep}{1.5pt}
    \begin{tabular*}{\textwidth}{
      @{\extracolsep{\fill}}lccc@{}
    }
      \toprule
      Method
        & MMLU-Pro
        & HellaSwag
        & \makecell{IFEval\\strict} \\
      \midrule

      \multicolumn{4}{@{}l}{\emph{WebShop, Qwen3-4B-Instruct}} \\
      Pretrained base
        & $65.25 \pm 0.66$
        & $80.14 \pm 0.07$
        & $83.30 \pm 0.56$ \\
      Full-history GRPO
        & $62.48 \pm 0.49\;(-2.77)$
        & $67.92 \pm 0.42\;(-12.22)$
        & $80.35 \pm 0.65\;(-2.96)$ \\
      CCM + GRPO
        & $55.16 \pm 0.60\;(-10.09)$
        & $66.68 \pm 0.71\;(-13.46)$
        & $82.32 \pm 0.47\;(-0.99)$ \\
      CCM + GRPO + distillation
        & $64.23 \pm 0.25\;(-1.03)$
        & $74.53 \pm 0.20\;(-5.62)$
        & $83.18 \pm 0.49\;(-0.12)$ \\

      \addlinespace
      \multicolumn{4}{@{}l}{\emph{WebShop, Qwen3-8B}} \\
      Pretrained base
        & $56.88 \pm 0.65$
        & $79.54 \pm 0.39$
        & $81.95 \pm 1.23$ \\
      Full-history GRPO
        & $54.28 \pm 1.00\;(-2.60)$
        & $76.68 \pm 0.80\;(-2.86)$
        & $79.48 \pm 1.40\;(-2.46)$ \\
      CCM + GRPO
        & $58.25 \pm 0.40\;(+1.37)$
        & $73.94 \pm 0.44\;(-5.60)$
        & $81.33 \pm 0.67\;(-0.62)$ \\
      CCM + GRPO + distillation
        & $56.95 \pm 0.25\;(+0.07)$
        & $76.71 \pm 0.17\;(-2.83)$
        & $81.58 \pm 1.17\;(-0.37)$ \\

\addlinespace
\multicolumn{4}{@{}l}{\emph{Endless Terminals, Qwen3-8B}} \\
Pretrained base
  & $56.88 \pm 0.65$
  & $79.54 \pm 0.39$
  & $81.95 \pm 1.23$ \\
Full-history GRPO
  & $58.63 \pm 0.80\;(+1.74)$
  & $79.31 \pm 0.42\;(-0.23)$
  & $82.44 \pm 0.67\;(+0.49)$ \\
CCM + GRPO
  & $60.61 \pm 0.92\;(+3.73)$
  & $77.38 \pm 0.12\;(-2.16)$
  & $82.44 \pm 1.33\;(+0.49)$ \\
CCM + GRPO + distillation
  & $61.75 \pm 0.37\;(+4.87)$
  & $78.48 \pm 0.79\;(-1.06)$
  & $82.38 \pm 0.56\;(+0.43)$ \\

      \bottomrule
    \end{tabular*}
  \end{table*}

On WebShop with Qwen3-4B-Instruct, privileged distillation substantially
reduces the degradation associated with CCM + GRPO. Relative to the
pretrained model, the distillation-augmented policy changes by
$-1.03$, $-5.62$, and $-0.12$ percentage points on MMLU-Pro, HellaSwag,
and IFEval. For WebShop with Qwen3-8B, the differences
between methods are smaller. Distillation improves HellaSwag and IFEval
retention relative to CCM + GRPO, while CCM + GRPO obtains the highest
MMLU-Pro score.

On Endless Terminals, all three post-trained policies improve over the
pretrained model on MMLU-Pro and remain close to it on IFEval. CCM + GRPO
+ distillation obtains the highest MMLU-Pro score and retains more
HellaSwag performance than CCM + GRPO, while full-history GRPO remains
closest to the pretrained HellaSwag score. Overall, privileged
distillation provides the clearest retention benefit for the 4B WebShop
model. Its effect is not uniform across model scales, training
environments, or evaluation tasks.

\section{Discussion and Conclusion}

CCM demonstrates that agents can operate with compact, continually updated memory, substantially reducing retained context. Privileged full-history distillation improves task accuracy while mitigating forgetting of general capabilities, with the clearest joint benefit on WebShop with Qwen3-4B-Instruct. This approach is particularly relevant when model weights and the KV cache compete for limited GPU memory, such as local agents running on consumer GPUs or servers supporting many concurrent agent sessions. Smaller active contexts could accommodate larger models or more simultaneous tasks within the same memory budget. However, the benefits vary across models and environments, and shorter prompts do not necessarily translate into lower API costs, which depend on the prefix caching and pricing policy.

\newpage
\subsection*{AI use statement}


Generative AI tools assisted with manuscript wording, LaTeX formatting,
and checks of bibliographic metadata against primary sources.

\subsection*{Ethics statement}

This work studies context management for agents that act in shopping and
terminal environments. Errors or omissions in retained memory can affect
later actions, and retained memory may contain sensitive information from
observations. Applications beyond these benchmarks should therefore use
appropriate access controls, data-handling safeguards, and human oversight
for consequential actions. The reported benchmark results do not establish
that CCM agents are safe or reliable for unrestricted deployment.

\subsection*{Reproducibility statement}

Section~\ref{sec:ccm} defines the CCM interaction protocol, and
Section~\ref{sec:method} describes the training objective.
Appendix~\ref{app:tb2-setup} provides the Terminal-Bench 2.0 inference
settings, prompt templates, and token accounting.
Appendices~\ref{app:grpo-objective} and~\ref{app:SDAR} specify the
optimization settings and distillation objective.
Appendix~\ref{app:experimental-details} documents the training and
evaluation splits, rewards, decoding settings, rollout limits, model
variants, and training framework; Appendix~\ref{app:ccm-prompts} provides
the CCM prompt templates. These details support reproduction of the
reported protocol, although results obtained through hosted model APIs
may vary with provider updates.




\bibliography{iclr2027_conference}
\bibliographystyle{iclr2027_conference}

\newpage
\appendix

\startcontents[appendix]
\section*{Appendix Contents}
\printcontents[appendix]{}{1}{\setcounter{tocdepth}{2}}
\clearpage

\section{Terminal-Bench 2.0 inference configuration}
\label{app:tb2-setup}

We evaluate pass@1 on the fixed 89-task Terminal-Bench 2.0 set. The
baseline uses the standard Terminus-2 agent and receives the complete
interaction history at every turn. Under CCM, each prompt instead contains
the original task, the memory written by the model on the preceding turn,
and the latest terminal observation. For all four models, we use
temperature $0.6$, allow at most 8{,}192 generated tokens per turn, and
limit each task to 50 turns. The CCM system prompt and per-turn input
template are shown in Listings~\ref{lst:tb2-ccm-system}
and~\ref{lst:tb2-ccm-turn}. The official task instruction is inserted
verbatim into the \texttt{\{instruction\}} field.

Table~\ref{tab:tb2-four-models-full} reports pass@1 performance and
cumulative token usage. Token counts are provider-reported means per task
and are summed across all model calls in an episode. Claude Sonnet 4.6,
Claude Opus 4.6, and Kimi K3 used provider-side prompt caching. GLM-5 was
evaluated without prompt caching because Amazon Bedrock did not support it
for this model.

\begin{table*}[t]
  \centering
  \caption{Detailed Terminal-Bench 2.0 pass@1 and mean cumulative token
  usage over 89 tasks. Baseline denotes full-history prompting.}
  \label{tab:tb2-four-models-full}

  \vskip 10pt
  \fontsize{8.5}{9.5}\selectfont
  \setlength{\tabcolsep}{1.5pt}
  \begin{tabular*}{\textwidth}{
    @{\extracolsep{\fill}}lcccccccc@{}
  }
    \toprule
    & \multicolumn{2}{c}{Claude Sonnet 4.6}
    & \multicolumn{2}{c}{Claude Opus 4.6}
    & \multicolumn{2}{c}{GLM-5}
    & \multicolumn{2}{c}{Kimi K3} \\
    \cmidrule(lr){2-3}
    \cmidrule(lr){4-5}
    \cmidrule(lr){6-7}
    \cmidrule(lr){8-9}
    Metric
      & Baseline & CCM
      & Baseline & CCM
      & Baseline & CCM
      & Baseline & CCM \\
    \midrule

    \multicolumn{9}{@{}l}{\emph{Evaluation}} \\
    Success, $n$ (\%)
      & \textbf{38 (42.7)} & 28 (31.5)
      & \textbf{52 (58.4)} & 36 (40.4)
      & \textbf{31 (34.8)} & 26 (29.2)
      & 56 (62.9) & \textbf{57 (64.0)} \\
    Average turns
      & 20.88 & 30.33
      & 21.60 & 26.00
      & 25.90 & 32.50
      & 14.49 & 19.02 \\

    \addlinespace
    \multicolumn{9}{@{}l}{\emph{Tokens per task}} \\
    Uncached input
      & 14{,}955 & 48{,}056
      & 23{,}914 & 86{,}052
      & 480{,}325 & 75{,}481
      & 102 & 27{,}039 \\
    Cache-read input
      & 408{,}900 & 27{,}135
      & 410{,}180 & 0
      & -- & --
      & 186{,}969 & 20{,}685 \\
    Cache-write input
      & 27{,}151 & 936
      & 30{,}440 & 0
      & -- & --
      & 14{,}776 & 1{,}128 \\
    Total input
      & 451{,}006 & 76{,}126
      & 464{,}533 & 86{,}052
      & 480{,}325 & 75{,}481
      & 201{,}846 & 48{,}852 \\
    Total output
      & 13{,}265 & 21{,}084
      & 12{,}662 & 26{,}286
      & 10{,}808 & 15{,}184
      & 12{,}659 & 35{,}354 \\
    Cache-read share (\%)
      & 90.7 & 35.6
      & 88.3 & 0.0
      & -- & --
      & 92.6 & 42.3 \\

    \bottomrule
  \end{tabular*}
\end{table*}

  \begin{lstlisting}[
    style=prompt,
    caption={CCM system prompt used for all four TerminalBench-2 models.},
    label={lst:tb2-ccm-system}
  ]
  You are an AI assistant tasked with solving command-line tasks in a Linux environment. You will be given a task description and
  the output from previously executed commands. Your goal is to solve the task by providing batches of shell commands.

  IMPORTANT MEMORY BEHAVIOR: You have a self-managed context window. You will not see the full previous trajectory. On each turn,
  you will receive the task description, your previous context_window, and the latest terminal output. Use context_window to
  preserve durable state needed for future turns.

  Format your response as JSON with the following structure:

  {
    "context_window": "Compact durable memory to carry into the next turn. Include task goal, working directory, key files
    inspected, important command results, edits made, tests run, unresolved blockers, and verification status. Do not include long
    terminal transcripts or private reasoning.",
    "analysis": "Analyze the current state based on the terminal output provided. What do you see? What has been accomplished? What
    still needs to be done?",
    "plan": "Describe your plan for the next steps. What commands will you run and why? Be specific about what you expect each
    command to accomplish.",
    "commands": [
      {
        "keystrokes": "ls -la\n",
        "duration": 0.1
      },
      {
        "keystrokes": "cd project\n",
        "duration": 0.1
      }
    ],
    "task_complete": true
  }

  Required fields:
  - "context_window": Compact durable memory to carry into the next turn. Include task goal, working directory, key files
  inspected, important command results, edits made, tests run, unresolved blockers, and verification status. Do not include long
  terminal transcripts, private reasoning, or speculative plans.
  - "analysis": Your analysis of the current situation
  - "plan": Your plan for the next steps
  - "commands": Array of command objects to execute

  Optional fields:
  - "task_complete": Boolean indicating if the task is complete

  Command object structure:
  - "keystrokes": String containing the exact keystrokes to send to the terminal (required)
  - "duration": Number of seconds to wait for the command to complete before the next command will be executed (defaults to 1.0 if
  not present)

  IMPORTANT: The text inside "keystrokes" will be used completely verbatim as keystrokes. Write commands exactly as you want them
  sent to the terminal:
  - You must end every command with a newline (\n) or it will not execute.
  - For special key sequences, use tmux-style escape sequences:
    - C-c for Ctrl+C
    - C-d for Ctrl+D

  The "duration" attribute specifies the number of seconds to wait for the command to complete (default: 1.0) before the next
  command will be executed. On immediate tasks (e.g., cd, ls, echo, cat) set a duration of 0.1 seconds. On commands (e.g., gcc,
  find, rustc) set a duration of 1.0 seconds. On slow commands (e.g., make, python3 [long running script], wget [file]) set an
  appropriate duration as you determine necessary.

  It is better to set a smaller duration than a longer duration. It is always possible to wait again if the prior output has not
  finished, by running {"keystrokes": "", "duration": 10.0} on subsequent requests to wait longer. Never wait longer than 60
  seconds; prefer to poll to see intermediate result status.

  Important notes:
  - Each command's keystrokes are sent exactly as written.
  - Do not include extra whitespace before or after the keystrokes unless it is part of the intended command.
  - Extra text before or after the JSON will generate warnings but be tolerated.
  - The JSON must be valid; use proper escaping for quotes and special characters within strings.
  - The commands array can be empty when waiting.
  \end{lstlisting}

  \begin{lstlisting}[
    style=prompt,
    caption={CCM per-turn input template. Braced fields are populated from the current task state.},
    label={lst:tb2-ccm-turn}
  ]
  Task Description:
  {instruction}

  Previous context_window:
  {context_window or "(empty)"}

  Current terminal state:
  Latest Terminal Output:
  {latest_terminal_output or "(none yet)"}

  Return the required JSON object for the next turn.
  \end{lstlisting}

\section{GRPO objective}
\label{app:grpo-objective}

For each task, GRPO samples a group of $G$ student trajectories and computes
group-normalized environment advantages,
\begin{equation}
  A^{(i)} = \frac{R^{(i)} - \mu_R}{\sigma_R + \varepsilon_{\mathrm{std}}}.
\end{equation}
Using the importance ratio
\begin{equation}
  r^{(i)}_t = \frac{\pi_\theta\bigl(y^{(i)}_t \mid s^{(i)}_t\bigr)}
                   {\pi_{\theta_{\mathrm{old}}}\bigl(y^{(i)}_t \mid s^{(i)}_t\bigr)},
\end{equation}
the clipped policy objective is
\begin{equation}
  L_{\mathrm{PG}} = -\operatorname{Agg}\Bigl[\min\bigl(r^{(i)}_t A^{(i)},\,
    \operatorname{clip}(r^{(i)}_t, 1-\epsilon, 1+\epsilon)\, A^{(i)}\bigr)\Bigr],
  \label{eq:pg}
\end{equation}
where $\operatorname{Agg}$ denotes the masked token mean over valid
student-generated tokens. Our implementation additionally uses dual clipping for
sufficiently negative advantages. A reference-policy regularizer is applied as
\begin{equation}
  L_{\mathrm{refKL}} = \operatorname{Agg}\Bigl[D_{\mathrm{KL}}\bigl(\pi_\theta(\cdot \mid s_t)
    \,\|\, \pi_{\mathrm{ref}}(\cdot \mid s_t)\bigr)\Bigr],
\end{equation}
implemented using a sampled KL estimator.

  \subsection{Optimization hyperparameters}

  Table~\ref{tab:training-hyperparameters} reports the shared optimization
  configuration. CCM + GRPO and CCM + GRPO + Distillation use identical GRPO
  hyperparameters. The latter additionally performs privileged full-history
  teacher scoring and applies the distillation loss. Its
  distillation-specific hyperparameters are
  $\lambda_{\mathrm{distill}}=0.01$ and $\beta=5$.

  \begin{table}[H]
    \centering
    \caption{Optimization hyperparameters used for WebShop and Endless
    Terminals.}
    \label{tab:training-hyperparameters}
    \vskip 10pt
    \fontsize{8.5}{9.5}\selectfont
    \setlength{\tabcolsep}{5pt}
    \begin{tabular}{@{}lcc@{}}
      \toprule
      Hyperparameter & WebShop & Endless Terminals \\
      \midrule
      Training steps                    & 100 & 60 \\
      Tasks per step                    & 32 & 32 \\
      Rollouts per task                 & 8 & 8 \\
      Trajectories per step             & 256 & 256 \\
      PPO mini-batch size               & 32 & 32 \\
      Micro-batch size per GPU          & 1 & 1 \\
      PPO epochs                        & 1 & 1 \\
      Learning rate                     & $1\times10^{-6}$ & $1\times10^{-6}$ \\
      LR schedule                       & Constant & Constant \\
      Warmup steps                      & 0 & 0 \\
      Adam betas                        & $(0.9,0.999)$ & $(0.9,0.999)$ \\
      Weight decay                      & 0.01 & 0.01 \\
      Gradient clipping                 & 1.0 & 1.0 \\
      PPO clip ratio                    & 0.2 & 0.2 \\
      Dual-clip coefficient             & 3.0 & 3.0 \\
      Entropy coefficient               & 0.001 & 0.0 \\
      Reference-KL coefficient          & 0.01 & 0.01 \\
      KL estimator                      & Low-variance & Low-variance \\
      Context-SDAR coefficient $\lambda$ & 0.01 & 0.01 \\
      Gate sharpness $\beta$            & 5.0 & 5.0 \\
      \bottomrule
    \end{tabular}
  \end{table}
 
\section{Details on the distillation objective}
\label{app:SDAR}

This section explains our choice of the SDAR \citep{lu2026sdar} objective,
derives its gradient, and establishes its connection to on-policy
distillation (OPD). Throughout, teacher probabilities are evaluated
on student-generated action tokens using privileged full-history
conditioning. The distillation objective supervises action tokens,
while GRPO provides task-level supervision for both memory and
action generation.

\paragraph{Motivation.}
The full interaction history provides information that may be
missing from the student's compact memory. Comparing the
probability of the same action token under these two contexts
therefore provides a training signal for behavior favored under
full-history conditioning. The teacher is a frozen copy of the student's
initial model, with weights held fixed throughout training; it is not
guaranteed to be stronger than the updated student.
Access to the full history does not guarantee that every teacher
judgment is correct, and large teacher--student probability gaps
need not indicate proportionally reliable supervision.

We therefore adopt the gap-gated distillation objective of
SDAR~\citep{lu2026sdar}. It converts the teacher--student
log-probability gap into a bounded gating weight, allowing
full-history guidance to modulate token-level likelihood gradients
while retaining GRPO as the task-level objective. The gate gives
greater weight to tokens assigned higher probability under
full-history conditioning and limits the gradient multiplier
associated with extreme probability disagreements. This choice
does not directly minimize reverse KL; its relationship to OPD
is established below.

\paragraph{Objective and implemented gradient.}
Let
\[
\ell^S_{i,n,t}(\theta)
=
\log \pi_\theta(a_{i,n,t}\mid s^C_{i,n,t}),
\qquad
\ell^T_{i,n,t}
=
\log \pi_{\theta_0}(a_{i,n,t}\mid s^+_{i,n,t}),
\]
where $\theta_0$ denotes the student's initial parameters before training.
The teacher weights are fixed at $\theta_0$ for the entire training run.
Student and teacher probabilities are evaluated
on the same sampled action token, conditioned on the compact
context and privileged full-history context, respectively.
Teacher scores are detached from gradient computation.

We define the detached log-probability gap and gating weight as
\begin{align}
\Delta_{i,n,t}
&=
\operatorname{sg}\!\left[
\ell^T_{i,n,t}-\ell^S_{i,n,t}(\theta)
\right],\\
g_{i,n,t}
&=
\operatorname{sg}\!\left[
\sigma(\beta\Delta_{i,n,t})
\right],
\end{align}
where $\operatorname{sg}$ denotes stop-gradient, $\sigma$ is the
logistic sigmoid, and $\beta>0$ controls gate sharpness.
Let $M_{i,n,t}$ indicate a valid action token with available
student and teacher scores, and let
$Z=\sum_{i,n,t}M_{i,n,t}$.
For a sampled batch with $Z>0$, the distillation loss is
\begin{equation}
L_{\mathrm{distill}}(\theta)
=
-\frac{1}{Z}
\sum_{i,n,t}
M_{i,n,t}g_{i,n,t}\ell^S_{i,n,t}(\theta).
\label{eq:appendix-distill-loss}
\end{equation}
Batches without eligible action tokens contribute no distillation
loss.

The sampled tokens, contexts, and masks are fixed during
backpropagation. Because the gating weights are detached,
\begin{equation}
\nabla_\theta L_{\mathrm{distill}}
=
-\frac{1}{Z}
\sum_{i,n,t}
M_{i,n,t}g_{i,n,t}
\nabla_\theta\ell^S_{i,n,t}(\theta).
\label{eq:appendix-distill-gradient}
\end{equation}
The teacher therefore affects the distillation gradient through
the gating weights, without requiring gradients through the
teacher model. Positive-gap tokens have $g_{i,n,t}>1/2$,
whereas negative-gap tokens have $g_{i,n,t}<1/2$.
Negative-gap tokens retain positive gating weights, scaling
down the magnitude of their individual distillation gradients
rather than reversing their likelihood-gradient direction.
These statements concern individual token gradients; the net
probability change also depends on other tokens and the GRPO
update through shared model parameters.

\paragraph{Equivalence to the SDAR loss.}
SDAR expresses its sampled-token objective using the weighted
teacher--student log-probability difference:
\begin{equation}
\widetilde L_{\mathrm{distill}}
=
\frac{1}{Z}
\sum_{i,n,t}
M_{i,n,t}g_{i,n,t}
\left[
\ell^T_{i,n,t}-\ell^S_{i,n,t}(\theta)
\right].
\end{equation}
This expression differs from
Eq.~(\ref{eq:appendix-distill-loss}) by
\[
C=
\frac{1}{Z}
\sum_{i,n,t}M_{i,n,t}g_{i,n,t}\ell^T_{i,n,t}.
\]
Since both the teacher scores and gating weights are detached,
$C$ has zero backpropagated gradient. The two expressions
therefore produce identical distillation gradients on the same
batch. We omit $C$ from the displayed objective to make the
weighted likelihood update explicit. This is a statement about
backpropagation at a given update: the numerical value of $C$
may change when scores or gates are recomputed.

\paragraph{Reverse-KL gradient in OPD.}
To compare the local token update with OPD, fix a student
context $s$ and its corresponding privileged context $s^+$.
Write
\[
p_\theta(a)=\pi_\theta(a\mid s),
\qquad
q(a)=\pi_{\theta_0}(a\mid s^+),
\qquad
\delta_\theta(a)=\log q(a)-\log p_\theta(a).
\]
Here $q$ is fixed, and both distributions are assumed positive
on the token support under consideration.

At these fixed contexts, the reverse-KL objective is
\begin{equation}
J_{\mathrm{RKL}}(\theta)
=
D_{\mathrm{KL}}(p_\theta\|q)
=
\sum_a p_\theta(a)
\left[\log p_\theta(a)-\log q(a)\right].
\end{equation}
Differentiating both the probability weights and the
log-probabilities gives
\begin{align}
\nabla_\theta J_{\mathrm{RKL}}
&=
\sum_a \nabla_\theta p_\theta(a)
\left[\log p_\theta(a)-\log q(a)+1\right]\\
&=
-\mathbb{E}_{a\sim p_\theta}
\left[
\delta_\theta(a)\nabla_\theta\log p_\theta(a)
\right].
\label{eq:appendix-opd-gradient}
\end{align}
The constant term vanishes because
\begin{equation}
\mathbb{E}_{a\sim p_\theta}
\left[\nabla_\theta\log p_\theta(a)\right]
=
\nabla_\theta\sum_a p_\theta(a)=0.
\label{eq:appendix-score-identity}
\end{equation}
Thus, OPD uses the signed log-probability gap as the coefficient
of the negative likelihood gradient.

\paragraph{Expected SDAR update and its connection to OPD.}
For a token $a$ sampled from the current student at the fixed
context, the backpropagated SDAR gradient is
\[
\widehat G_{\mathrm{SDAR}}(\theta;a)
=
-\sigma(\beta\delta_\theta(a))
\nabla_\theta\log p_\theta(a).
\]
Define its expectation as
\begin{equation}
G_{\mathrm{SDAR}}(\theta)
=
\mathbb{E}_{a\sim p_\theta}
\left[\widehat G_{\mathrm{SDAR}}(\theta;a)\right].
\end{equation}
This is the expectation of the implemented sampled-token
gradient. It should not be confused with fully differentiating
a current-policy expectation of the sampled-token loss, which
would introduce additional sampling-distribution terms.

Using Eq.~(\ref{eq:appendix-score-identity}), we can subtract
$1/2$ from the gating weight without changing the expected
gradient:
\begin{align}
G_{\mathrm{SDAR}}
&=
-\mathbb{E}_{a\sim p_\theta}
\left[
\left(\sigma(\beta\delta_\theta(a))-\frac12\right)
\nabla_\theta\log p_\theta(a)
\right]\\
&=
-\frac12\mathbb{E}_{a\sim p_\theta}
\left[
\tanh\!\left(\frac{\beta\delta_\theta(a)}{2}\right)
\nabla_\theta\log p_\theta(a)
\right].
\label{eq:appendix-centered-gradient}
\end{align}

Consequently, the effective token weights in the expected
OPD and SDAR gradients are
\begin{equation}
w_{\mathrm{OPD}}(\delta)=\delta,
\qquad
w_{\mathrm{SDAR}}(\delta)
=
\sigma(\beta\delta)-\frac12
=
\frac12\tanh(\beta\delta/2).
\end{equation}
The centered SDAR weight is monotone, has the same sign as
$\delta$, and is bounded between $-1/2$ and $1/2$.
It therefore preserves the sign and ordering of token-level
teacher preferences while compressing their magnitudes.
Although the implemented gate is positive, its expected
on-policy gradient admits this equivalent signed
representation.

For $|\beta\delta|\ll1$, a Taylor expansion gives
\begin{equation}
w_{\mathrm{SDAR}}(\delta)
=
\frac{\beta}{4}\delta
+O\!\left((\beta\delta)^3\right).
\end{equation}
When this approximation holds for the tokens contributing
to the expectation,
\begin{equation}
G_{\mathrm{SDAR}}(\theta)
\approx
\frac{\beta}{4}
\nabla_\theta D_{\mathrm{KL}}(p_\theta\|q).
\label{eq:appendix-local-opd}
\end{equation}
The expected SDAR update thus approaches a scaled OPD gradient
locally for sufficiently small log-probability gaps. For large
gaps, the effective weight saturates rather than growing
linearly. Because this transformation reweights tokens
nonlinearly, the aggregate SDAR and OPD gradients need not
have the same direction away from the small-gap regime.

\paragraph{Scope of the connection.}
The centering identity and local approximation characterize
a token update at fixed contexts under sampling from the
current student distribution. They do not differentiate
through the generation of memories, preceding action tokens,
or the distribution of visited environment histories.
They also do not establish exact reverse-KL minimization
for the complete training procedure.

In practice, rollouts are generated by a behavior-policy
snapshot and may be reused after parameter updates.
The exact on-policy identity then need not hold without
appropriate sampling corrections. Token-dependent filtering
and normalization by a random number of eligible tokens can
also change the expected batch gradient. The derivation therefore
characterizes the distillation update locally; it does not establish an
exact identity for the full masked trajectory loss.

Finally, subtracting $1/2$ preserves the expected gradient
under the stated assumptions, but generally changes a
finite-batch gradient. Our implementation retains the
positive sigmoid gate; centering is used only for analysis.
The bounded gate limits the per-token gradient multiplier,
not the norm of the underlying model gradient, and does
not guarantee monotonic KL reduction or task improvement.

  \section{Additional Ablation Results}
  \label{app:ablations}

  \subsection{Distillation without GRPO}
  \label{app:distillation-only}

  We evaluate whether auxiliary privileged distillation can train a CCM
  policy without the accompanying GRPO objective. These runs retain
  same-trajectory privileged full-history scoring on student action tokens
  but set the GRPO loss weight to zero. Because the distillation objective
  operates only on action tokens, memory tokens receive no direct training
  signal in this condition.

  \begin{table}[t]
    \centering
    \fontsize{8.5}{9.5}\selectfont
    \setlength{\tabcolsep}{3pt}
    \caption{Distillation-only ablation on the fixed 128-task WebShop
    evaluation. Both models are evaluated at checkpoint 100 under CCM.
    Partial score is the mean percentage of WebShop constraints satisfied.}
    \label{tab:webshop-distillation-only}
    \vskip 10pt
    \begin{tabular}{@{}lcccc@{}}
      \toprule
      Model & Success (\%) & Solved & Partial score (\%) & Turns \\
      \midrule
      Qwen3-4B-Instruct & 0.00 & 0/128 & 0.59 & 14.97 \\
      Qwen3-8B          & 0.78 & 1/128 & 5.27 & 14.11 \\
      \bottomrule
    \end{tabular}
  \end{table}

  As shown in Table~\ref{tab:webshop-distillation-only}, both
  distillation-only policies achieve nearly zero exact success at checkpoint
  100: Qwen3-4B-Instruct solves no tasks, while Qwen3-8B solves only one of
  128. In these WebShop experiments, privileged action-token distillation
  therefore does not replace verifier-driven GRPO. GRPO provides
  trajectory-level supervision to both memory and action tokens, while
  privileged distillation is effective only as an auxiliary objective
  alongside GRPO in the main experiments.

\section{Experimental Details}
  \label{app:experimental-details}

  \subsection{Benchmarks and data splits}

  \paragraph{WebShop.}
  We use the WebShop text environment with the small product configuration.
Training tasks are sampled from all WebShop task IDs greater than or equal
  to 500. Evaluation uses the fixed set of task IDs 0--127. We train for 100 policy updates and evaluate
  saved checkpoints separately using pass@1, temperature $0.4$, seed $42$, and a
  15-turn episode limit.

  We use a shaped reward that combines WebShop's partial task score with a
  larger bonus for exact task completion. Let
  $r_{\mathrm{partial}}\in[0,1]$ denote the native WebShop task score and
  let $s\in\{0,1\}$ indicate exact success. The training reward is
  \begin{equation}
    R_{\mathrm{WebShop}}
    =
    2r_{\mathrm{partial}} + 8s.
    \label{eq:webshop-training-reward}
  \end{equation}
  Thus, unsuccessful trajectories receive rewards between $0$ and $2$ according
  to their partial progress, whereas an exactly successful trajectory receives
  reward $10$.

  \paragraph{Endless Terminals.}
  We use the fixed Endless Terminals split containing 2,083 training tasks,
  100 validation tasks, and 300 held-out test tasks. Training lasts for 60
  policy updates. Final results are obtained by evaluating saved checkpoints on
  the fixed 300-task test set with pass@1, seed $42$, temperature $0.6$, and a
  16-turn episode limit. The environment reward is binary task success. Invalid
  actions receive an auxiliary penalty with coefficient $0.1$.

  \subsection{Rollout configuration}

  Table~\ref{tab:rollout-hyperparameters} summarizes the rollout and sequence
  limits. Thinking is disabled for all reported experiments. The self-managed
  student emits one memory block followed by one environment-action block at
  every turn.

  \begin{table}[H]
    \centering
    \caption{Rollout and sequence-length configuration. WebShop uses 1,024
    generation tokens for Qwen3-4B-Instruct and 2,048 for Qwen3-8B.}
    \label{tab:rollout-hyperparameters}
    \vskip 10pt
    \fontsize{8.5}{9.5}\selectfont
    \setlength{\tabcolsep}{5pt}
    \begin{tabular}{@{}lcc@{}}
      \toprule
      Hyperparameter & WebShop & Endless Terminals \\
      \midrule
      Maximum turns                  & 15 & 16 \\
      Sampling temperature           & 1.0 & 0.6 \\
      Top-$p$                        & 1.0 & 1.0 \\
      Top-$k$                        & Unrestricted & Unrestricted \\
      Student prompt limit           & 4,096 & 16,384 \\
      Teacher prompt limit           & 32,768 & 16,384 \\
      Generation tokens per turn     & 1,024 / 2,048 & 4,096 \\
      Maximum model length           & 32,768 & 20,480 \\
      Thinking enabled               & No & No \\
      \bottomrule
    \end{tabular}
  \end{table}

  \subsection{Models and systems}

  WebShop experiments use Qwen3-4B-Instruct-2507 and Qwen3-8B. Endless
  Terminals experiments use Qwen3-8B. Training is implemented in \textsc{verl}
  with FSDP policy updates and colocated vLLM rollout generation. Each run uses
  one node with four H100 or H200 GPUs.

\subsection{Evaluation}
 WebShop checkpoint evaluation uses 
  temperature $0.4$, seed $42$, and the fixed 128-task set. Endless Terminals
use temperature $0.6$, seed $42$, and
  the fixed 300-task held-out test set.

\subsection{CCM Prompt Templates}
  \label{app:ccm-prompts}

  The following listings reproduce the prompts used for Continuous Context
  Management. WebShop uses a single per-turn prompt containing the task,
  retained memory, current observation, and admissible actions. Endless
  Terminals uses a fixed system prompt, followed by the original task as a
  separate user message and a per-turn user message containing the retained
  memory and newest terminal observation. Braced fields are populated at
  runtime.

  \begin{lstlisting}[
    basicstyle=\ttfamily\scriptsize,
    breaklines=true,
    columns=fullflexible,
    keepspaces=true,
    showstringspaces=false,
    caption={WebShop CCM per-turn prompt template.},
    label={lst:webshop-ccm-prompt}
  ]
  You are an expert autonomous agent operating in the WebShop e-commerce environment.
  Your task is to: {task_description}.

  MEMORY:
  - Memory is the persistent record of useful information from earlier shopping
    turns. The original task is always shown separately.
  - Your previous accepted memory appears below. It is empty on the first turn.
  - Every response must contain exactly one <memory> </memory> block.
  - Inside that block, retain concrete task constraints, product evidence,
    selected options, and unresolved requirements that could be needed later.
  - If no useful information changed, repeat the previous memory unchanged.
  - Never write instructions, labels, or generic filler in memory.

  Previous memory:
  <memory>{context_window}</memory>

  You are now at step {current_step}.
  Your current observation is: {current_observation}.
  Your admissible actions are:
  [
  {available_actions}
  ].

  Write the updated <memory> </memory> block. Then choose exactly one admissible
  action and write it inside <action> </action>.
  \end{lstlisting}

  \begin{lstlisting}[
    basicstyle=\ttfamily\scriptsize,
    breaklines=true,
    columns=fullflexible,
    keepspaces=true,
    showstringspaces=false,
    caption={Endless Terminals CCM system prompt.},
    label={lst:endless-ccm-system}
  ]
  You are a highly capable Linux terminal agent operating strictly via a single-shell-command interface.
  Goal: complete the user's task.

  CRITICAL MEMORY REQUIREMENT:
  - EVERY RESPONSE MUST BEGIN WITH <memory> AND INCLUDE EXACTLY ONE </memory> CLOSING TAG.
  - THIS INCLUDES THE FIRST TURN, EVERY INTERMEDIATE TURN, AND THE FINAL TURN.
  - A RESPONSE CONTAINING <command> OR <action>done</action> WITHOUT <memory>...</memory> IS INVALID AND WILL NOT EXECUTE.
  - NEVER OUTPUT A BARE COMMAND. YOUR FIRST OUTPUT CHARACTERS MUST BE: <memory>

  Rules:
  - Every response must use exactly one of these two forms:

    <memory>FACTS_TO_REMEMBER</memory>
    <command>THE_SINGLE_SHELL_COMMAND</command>

    OR

    <memory>FACTS_TO_REMEMBER</memory>
    <action>done</action>

  - Do not output a bare <command> or <action> block.
  - Do not use <command_window> or any other wrapper names.
  - Don't use interactive commands and confirmations; use non-interactive flags.
  - Prefer simple, robust CLI tools; write files explicitly when needed.
  - If you believe the task is solved, use the second form with <action>done</action>.
  - Hint: you might want to run commands interactively to see the output and then write the command. Don't just pipe the commands.
  - Only your first command in command tags will be executed. So don't respond with multiple commands.
  - Hint: Verify your solution once you are done. Eg: you can use cat to see the input and the output.
  - Do not just write long bash scripts. Write the commands that you would write in a terminal.

  IMPORTANT MEMORY BEHAVIOR:
  - Write in <memory> the facts from the task and terminal work that should be available on the next turn.
  - The next turn will not receive the earlier conversation. It receives only the original task, your latest memory, and the newest terminal
  observation.
  - Update the memory as work progresses so it reflects the current durable state.
  - Memory describes the state before the command in the same response executes.
  - Record completed work only when confirmed by an earlier terminal observation.
  - Do not describe the accompanying command as already completed.
  \end{lstlisting}

  In Endless Terminals, the original task instruction is supplied verbatim as
  a separate user message. The following additional user message is then
  reconstructed at every turn:

  \begin{lstlisting}[
    basicstyle=\ttfamily\scriptsize,
    breaklines=true,
    columns=fullflexible,
    keepspaces=true,
    showstringspaces=false,
    caption={Endless Terminals CCM per-turn user-message template.},
    label={lst:endless-ccm-turn}
  ]
  Previous memory:
  <memory>{memory}</memory>

  Current observation:
  {observation}

  RESPONSE REQUIREMENT: Begin your response with <memory>. Write the complete updated <memory>...</memory> block, then exactly one <command>...</
  command> or <action>done</action>. Never return a bare command.
  \end{lstlisting}


\section{Qualitative Trajectory Analysis}
\label{app:qualitative-traces}

We present condensed matched trajectories to illustrate behavioral differences
among the evaluated conditions. For each trajectory, we retain the initial
turn, turns that materially change the retained memory or environment state,
the first instance of each repeated failure mode, the final effective action,
and the verifier outcome. Consecutive turns with equivalent behavior are
summarized explicitly. Bracketed omissions inside excerpts are editorial and
remove only irrelevant or repeated output. 

\subsection{WebShop}
\label{app:qualitative-webshop}

\paragraph{Task.}
The agent must purchase a men's shirt priced below \$50 that satisfies the
requested material, fit, sleeve, color, and size constraints. The matched task
requires cotton spandex, classic fit, short sleeves, charcoal heather (light),
and size 1x.

\begin{table*}[t]
  \centering
  \caption{Outcomes for the matched WebShop trajectory. Exact denotes exact
  task success; partial is the WebShop constraint-matching score.}
  \label{tab:qualitative-webshop-outcomes}
  \vskip 10pt
  \fontsize{8.5}{9.5}\selectfont
  \setlength{\tabcolsep}{6pt}
  \begin{tabular*}{\textwidth}{@{\extracolsep{\fill}}llrrr@{}}
    \toprule
    Model & Condition & Exact & Partial & Turns \\
    \midrule
    Qwen3-4B-Instruct & Untrained full history        & No  & 0.000 & 15 \\
                      & Untrained CCM                 & No  & 0.000 & 15 \\
                      & CCM + GRPO                    & No  & 0.857 & 15 \\
                      & CCM + GRPO + Distillation     & Yes & 1.000 & 5  \\
    \addlinespace
    Qwen3-8B          & Untrained full history        & No  & 0.000 & 12 \\
                      & Untrained CCM                 & No  & 0.000 & 15 \\
                      & CCM + GRPO                    & No  & 0.857 & 10 \\
                      & CCM + GRPO + Distillation     & Yes & 1.000 & 5  \\
    \bottomrule
  \end{tabular*}
\end{table*}

\paragraph{Untrained policies.}
The untrained full-history policies repeatedly issue broad searches without
reaching a valid purchase. The untrained CCM policies preserve the task
constraints in memory but continue paging through irrelevant results. Neither
condition purchases a product satisfying the requested options.

\paragraph{CCM + GRPO failure.}
For Qwen3-4B-Instruct, CCM + GRPO finds the target product and opens its product
page. Its memory, however, remains effectively unchanged after the initial
search:

\begin{Verbatim}[breaklines=true,breakanywhere=true,fontsize=\scriptsize]
<memory>
Task: Find machine wash men's dress shirts with cotton spandex,
classic fit, short sleeve, color: charcoal heather (light),
size: 1x, price lower than $50.00.
Unresolved: No search or product options have been generated yet.
</memory>
<action>click[b07hrfsnl4]</action>
\end{Verbatim}

The resulting product page exposes both required options:

\begin{Verbatim}[breaklines=true,breakanywhere=true,fontsize=\scriptsize]
size: [..., 1x, ...]
color: [..., charcoal heather (light), ...]
Nautica Men's Solid Crew Neck Short-Sleeve Pocket T-Shirt
Price: $16.05 to $40.98
Buy Now
\end{Verbatim}

The policy selects the color but then repeats the same color-selection action
through the remaining intermediate turns instead of selecting size 1x:

\begin{Verbatim}[breaklines=true,breakanywhere=true,fontsize=\scriptsize]
<action>click[charcoal heather (light)]</action>
\end{Verbatim}

At the final turn it purchases the product with only the color selected. The
environment assigns partial score $0.857$ rather than exact success:

\begin{Verbatim}[breaklines=true,breakanywhere=true,fontsize=\scriptsize]
Purchased
asin: B07HRFSNL4
options: {"color": "charcoal heather (light)"}
Reward: 0.8571428571428571
\end{Verbatim}

The Qwen3-8B CCM + GRPO trajectory exhibits the same high-level failure: it
reaches the target product but purchases without satisfying every required
option, receiving partial score $0.857$.

\paragraph{CCM + GRPO + Distillation success.}
The distillation-augmented Qwen3-4B-Instruct policy records the task constraints
and the current stage of the interaction. Its five actions form the required
search--product--color--size--purchase sequence:

\begin{Verbatim}[breaklines=true,breakanywhere=true,fontsize=\scriptsize]
Turn 1: <action>search[cotton men's dress shirts cotton spandex
        classic fit short sleeve color charcoal heather (light)
        size 1x price under 50.00]</action>
Turn 2: <action>click[b07hrfsnl4]</action>
Turn 3: <action>click[charcoal heather (light)]</action>
Turn 4: <action>click[1x]</action>
Turn 5: <action>click[buy now]</action>
\end{Verbatim}

The memory changes with the option-selection state. For example, immediately
before the size action it records:

\begin{Verbatim}[breaklines=true,breakanywhere=true,fontsize=\scriptsize]
<memory>
Task: Find machine wash men's dress shirts with cotton spandex,
classic fit, short sleeve, color: charcoal heather (light),
size: 1x, price lower than $50.00.
Current step: 4 (Select size).
Action: Select size 1x.
Size selected: 1x.
Size constraint met: yes.
</memory>
<action>click[1x]</action>
\end{Verbatim}

The final observation confirms that both required options were selected:

\begin{Verbatim}[breaklines=true,breakanywhere=true,fontsize=\scriptsize]
Purchased
asin: B07HRFSNL4
options: {"color": "charcoal heather (light)", "size": "1x"}
Reward: 1.0
\end{Verbatim}

Qwen3-8B follows the same five-action sequence. Its memory uses a structured
representation that retains the constraints and selected actions:

\begin{Verbatim}[breaklines=true,breakanywhere=true,fontsize=\scriptsize]
<memory>
{
  "constraints": {
    "type": "men's dress shirts",
    "material": "cotton spandex",
    "fit": "classic fit",
    "sleeve": "short sleeve",
    "color": "charcoal heather (light)",
    "size": "1x",
    "price_limit": 50.00
  },
  "actions_taken": [
    "click[charcoal heather (light)]",
    "click[1x]"
  ]
}
</memory>
\end{Verbatim}

It likewise purchases the target product with both options selected and
receives exact reward 1.0.

\subsection{Endless Terminals}
\label{app:qualitative-endless}

The two Endless Terminals examples expose complementary outcomes: one is solved
only by CCM + GRPO + Distillation among the matched conditions, while the other
is solved by CCM + GRPO but not by the distillation-augmented policy. They also
contrast clean early termination with continued action after a verifier-valid
state has been created.

\begin{table*}[t]
  \centering
  \caption{Outcomes for the two matched Endless Terminals trajectories.}
  \label{tab:qualitative-endless-outcomes}
  \vskip 10pt
  \fontsize{8.5}{9.5}\selectfont
  \setlength{\tabcolsep}{6pt}
  \begin{tabular*}{\textwidth}{@{\extracolsep{\fill}}llrr@{}}
    \toprule
    Task & Condition & Success & Turns \\
    \midrule
    \texttt{4d41da7b} & Untrained full history    & No  & 2  \\
                       & Untrained CCM             & No  & 2  \\
                       & CCM + GRPO                & No  & 16 \\
                       & CCM + GRPO + Distillation & Yes & 2  \\
    \addlinespace
    \texttt{beff73f4} & Untrained full history    & Yes & 4  \\
                       & Untrained CCM             & No  & 16 \\
                       & CCM + GRPO                & Yes & 16 \\
                       & CCM + GRPO + Distillation & No  & 16 \\
    \bottomrule
  \end{tabular*}
\end{table*}

\subsubsection{Permission-audit setup: distillation-augmented CCM success}

\paragraph{Task.}
The agent must create a shell-script stub, set directory and file modes to 750
and 740, respectively, and write a three-line permission-verification log with
exact contents.

\paragraph{Failure modes.}
The untrained full-history policy performs nearly all required operations, but
constructs the third log line from \texttt{ls -l}. The resulting permission
string contains a trailing SELinux marker, \texttt{-rwxr-----.}, rather than the
required \texttt{-rwxr-----}. The untrained CCM policy creates only the
directory, then records in memory that the unexecuted file and log operations
have already been completed and terminates. CCM + GRPO initially creates a
valid state, but its memory remains unchanged:

\begin{Verbatim}[breaklines=true,breakanywhere=true,fontsize=\scriptsize]
<memory>Initial state: No commands have been executed yet.</memory>
\end{Verbatim}

It continues issuing commands through turn 16 and ultimately overwrites the
verification log with only its third line. The verifier therefore reports that
the file contains one line rather than three.

\paragraph{CCM + GRPO + Distillation.}
The distillation-augmented policy performs the complete setup in one command:

\begin{Verbatim}[breaklines=true,breakanywhere=true,fontsize=\scriptsize]
<memory>Need to create a specific directory structure and files
with specific permissions and content.</memory>
<command>mkdir -p /home/user/project_utils/secure_scripts &&
chmod 750 /home/user/project_utils/secure_scripts &&
echo -e '#!/bin/bash\necho "Backup script stub"' >
/home/user/project_utils/secure_scripts/backup.sh &&
chmod 740 /home/user/project_utils/secure_scripts/backup.sh &&
echo -e "$(stat -c "%a %n" /home/user/project_utils/secure_scripts)\n\
$(stat -c "%a %n" /home/user/project_utils/secure_scripts/backup.sh)\n\
-rwxr----- /home/user/project_utils/secure_scripts/backup.sh" >
/home/user/perm_check.log</command>
\end{Verbatim}

After observing a zero exit code, it requests completion on turn 2:

\begin{Verbatim}[breaklines=true,breakanywhere=true,fontsize=\scriptsize]
<memory>Need to create a specific directory structure and files
with specific permissions and content.</memory>
<action>done</action>

... [100%]
3 passed in 0.01s
\end{Verbatim}

This trajectory provides a clean early-termination example: the policy reaches
the verifier-valid state with one environment action and does not issue any
redundant commands.

\subsubsection{Git tracked-file export: CCM + GRPO success}

\paragraph{Task.}
The agent must query Git inside \texttt{/home/user/demo\_repo} and write all
tracked paths, in Git's returned order and with a final newline, to
\texttt{/home/user/git\_files.log}.

\paragraph{Untrained full-history baseline.}
After two commands fail because they run outside the repository, the baseline
changes directory and succeeds:

\begin{Verbatim}[breaklines=true,breakanywhere=true,fontsize=\scriptsize]
<command>cd /home/user/demo_repo && git ls-files |
tee /home/user/git_files.log</command>

README.md
scripts/run.sh
src/main.py
\end{Verbatim}

It requests completion on turn 4, and all five verifier tests pass.

\paragraph{CCM + GRPO.}
The first CCM + GRPO command also runs in the wrong directory and fails. The
second turn corrects the working directory and queries Git:

\begin{Verbatim}[breaklines=true,breakanywhere=true,fontsize=\scriptsize]
<memory>
Initial state: No commands have been executed yet.
Task: List tracked files in /home/user/demo_repo and save to
/home/user/git_files.log.
</memory>
<command>cd /home/user/demo_repo && git ls-files</command>

Command succeeded. Output:
README.md
scripts/run.sh
src/main.py
\end{Verbatim}

The following memory preserves the command result, and the policy writes the
three paths to the requested file:

\begin{Verbatim}[breaklines=true,breakanywhere=true,fontsize=\scriptsize]
<memory>
Task: List tracked files in /home/user/demo_repo and save to
/home/user/git_files.log.
Output of git ls-files: README.md scripts/run.sh src/main.py
</memory>
<command>echo -e "README.md\nscripts/run.sh\nsrc/main.py" >
/home/user/git_files.log</command>
\end{Verbatim}

Turns 4--16 repeatedly rewrite the same valid content instead of terminating.
The final state passes all verifier tests:

\begin{Verbatim}[breaklines=true,breakanywhere=true,fontsize=\scriptsize]
..... [100%]
5 passed in 0.02s
\end{Verbatim}

\paragraph{CCM + GRPO + Distillation failure.}
The distillation-augmented policy initially mistakes a wrong working directory
for an absent repository and unnecessarily reinitializes Git. It then repeatedly
uses the invalid \texttt{-n} option with \texttt{git ls-files}. Although turn
15 finally obtains the correct tracked paths, turn 16 again uses the invalid
option and leaves an incorrect output file. The verifier reports three failed
tests, including incorrect contents and a missing trailing newline.

\paragraph{Interpretation.}
The two examples show that neither trained CCM objective dominates on every
task. They also show that termination behavior varies substantially across
conditions: the distillation-augmented policy terminates immediately after its
successful action in the permission task, whereas CCM + GRPO establishes a
correct state early in the Git task but continues issuing redundant commands
until the turn limit.

\end{document}